%% file: root.tex
\documentclass[letterpaper, 10 pt, conference]{ieeeconf}  

\usepackage{tikz}
\usetikzlibrary{snakes}
\usetikzlibrary{positioning}
\usepackage{pgfplots}
\usepackage{pgfplotstable}
\pgfplotsset{compat=1.7}
 
\usepgfplotslibrary{groupplots}
\usepackage{subcaption}
\usepackage{url}
\let\labelindent\relax
\usepackage{enumitem}
\usepackage{balance} 
\AtBeginDocument{%
  \providecommand\BibTeX{{%
    \normalfont B\kern-0.5em{\scshape i\kern-0.25em b}\kern-0.8em\TeX}}}

\IEEEoverridecommandlockouts                              

\title{\LARGE \bf
Ice-Breakers, Turn-Takers and Fun-Makers: Exploring Robots for Groups with Teenagers
}

\author{Sarah Gillet*$^{1}$, Katie Winkle*$^{1}$, Giulia Belgiovine*$^{2}$ and Iolanda Leite$^{1}$ 
\thanks{*Authors contributed equally.}
\thanks{$^{1}$Sarah Gillet, Katie Winkle, and Iolanda Leite are with the Division of Robotics, Perception and Learning, School of Electrical Engineering and Computer Science,
       KTH Royal Institute of Technology, Sweden
        {\tt\small \{winkle, sgillet, iolanda\}@kth.se}}%
\thanks{$^{2}$Giulia Belgiovine is with the CONTACT unit at Istituto Italiano di Tecnologia and DIBRIS department of Università degli Studi di Genova 
{\tt\small giulia.belgiovine@iit.it}}%
\thanks{This work was partially funded by the Swedish Research Council (no. 2017-05189), the Swedish Foundation for Strategic Research (FFL18-0199), the Jacobs Foundation (no. 2017 1261 06), Digital Futures, and the Vinnova Competence Center for Trustworthy Edge Computing Systems and Applications at KTH. 
}
}

\begin{document}

\maketitle
\thispagestyle{empty}
\pagestyle{empty}

\input{sections/abstract}

\input{sections/01_introduction}
\input{sections/02_relatedwork}

\input{sections/03_hypothesis}
\input{sections/04_envisioning_robot_use}
\input{sections/06_discussion}
\input{sections/07_conclusion}

\bibliographystyle{IEEEtran}
\balance
\bibliography{MyLibrary,bibliography}

\end{document}

%% file: sections/abstract.tex
\begin{abstract}
Successful, enjoyable group interactions are important in public and personal contexts, especially for teenagers whose peer groups are important for self-identity and self-esteem. Social robots seemingly have the potential to positively shape group interactions, but it seems difficult to effect such impact by designing robot behaviors solely based on related (human interaction) literature. 
In this article, we 
take a user-centered approach to explore how teenagers envisage a social robot ``group assistant''. We engaged 16 teenagers in focus groups, interviews, and robot testing to capture their views and reflections about robots for groups. Over the course of a two-week summer school, participants co-designed the action space for such a robot and experienced working with/wizarding it for 10+ hours. This experience further altered and deepened their insights into using robots as group assistants. 
We report results regarding teenagers' views on the applicability and use of a robot group assistant, how these expectations evolved throughout the study, and their repeat interactions with the robot. Our results indicate that each group moves on a spectrum of need for the robot, reflected in use of the robot more (or less) for ice-breaking, turn-taking, and fun-making as the situation demanded. 
\end{abstract}

%% file: sections/01_introduction.tex
\section{Introduction}
Interacting in groups is an essential element of everyday human life. Especially for teenagers, peer groups are important for self-identity and self-esteem~\cite{brown1987peer}.
Essential to a group's function and the behaviour of its members are the group dynamics, such as cohesion. For example, among teenagers, higher cohesion has been found to lead to more generalist trust and more prosocial behaviours~\cite{bosSocialNetworkCohesion2018}. 

Recent works have demonstrated how social robots can interact in groups~\cite{Sebo2020RobotsReview}. Specifically, a number of works investigated how robots might assist a group to improve the group dynamics, e.g. help in situations of conflict~\cite{Martelaro2015}, balance engagement and participation in conversations~\cite{Tennent2019, RobotLevelGilletCumbal2021}, or shape the perception of cohesion among children~\cite{Strohkorb2016}. 


These prior efforts often leverage known phenomena from psychology or social science literature to hand-craft the robot's behaviours. However, prior works also report that these literature-informed behaviours sometimes generate different effects than intended~\cite{Strohkorb2016, Martelaro2015, Short2017Understanding}.
Motivated by these findings, we take a participatory design (PD) and user-centred approach to explore how teenagers envisage a social robot which aims to improve their group interactions. 
Such approaches have been used successfully in the design of social (and particularly socially \textit{assistive} robots) e.g. for use in healthcare~\cite{leeStepsParticipatoryDesign2017} and education~\cite{alves-oliveiraYOLORobotCreativity2017}. To our knowledge, no previous works have applied such an approach specifically to developing robots for groups.

To involve teenagers in the design of social robot behaviours, we combined a methodology for expert-led design and automation of social robots (previously used exclusively with adult experts creating dyadic interactions~\cite{winkle2021leador}) with best practices for working with teenagers~\cite{bjorling2019participatory,bjorling2020exploring}. 
As a result, we (1) engaged teenagers in focus groups and interviews to discuss robots in groups, (2) involved them in creating a discrete set of robot actions designed to support group interactions, and (3) invited them to wizard and experience those jointly designed actions across multiple, in-situ group-robot interactions. 




\begin{figure}[t]
\centerline{\includegraphics[width=0.91\linewidth]{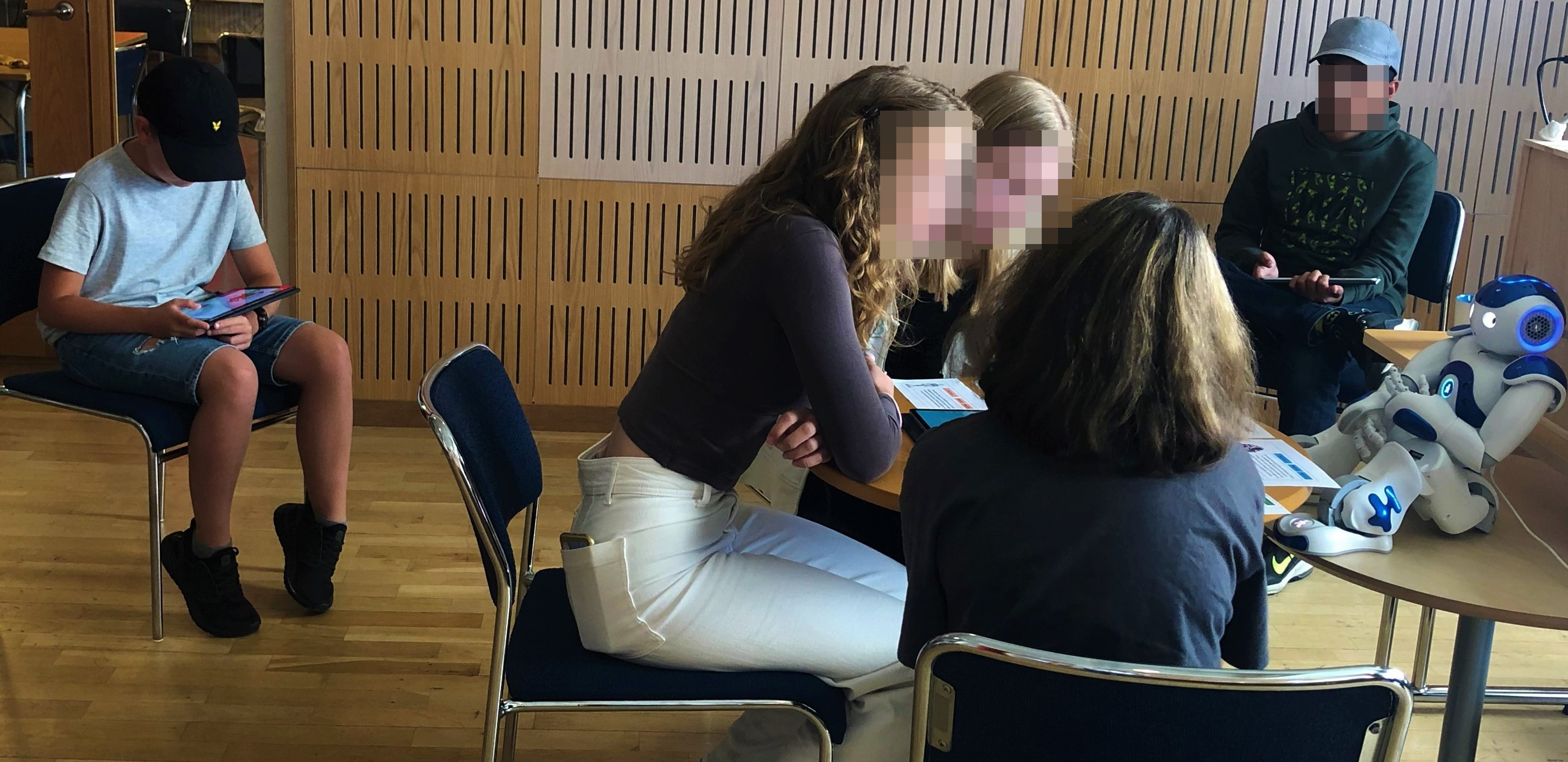}}
\caption{The group, robot and teaching setup: teen left is controlling the robot's actions (Robot Controller, RC); teen right is reporting on group behavior (Group Observer, GO); the other three teens are working on a discussion-based group activity assisted by the robot (Group Members, GM).}
\label{fig:training}
\end{figure}


Drawing qualitative insights from our PD activities and the resultant in-situ testing with teenagers, this article discusses what role a robot ``group assistant'' could have and how this role might need to change over time and/or between groups. 
In this way, we contribute to deepening state of the art understanding on the role/impact of social robots in (teenagers') groups. 

%% file: sections/02_relatedwork.tex
\section{Related Work}

\subsection{Robots in Groups}
The study of robots and groups has gained importance within the field of HRI~\cite{NonDyadicSchneiders22}, 
including how people perceive robots in groups and how they influence and facilitate group dynamics~\cite{Sebo2020RobotsReview}. 
In particular, robots have been shown to improve situations of conflict \cite{Martelaro2015, Shen2018} and emotional support \cite{erel2021enhancing} and foster the expression of vulnerability \cite{StrohkorbSebo2018} or perception of cohesion~\cite{Strohkorb2016}. Further, prior work has been interested in studying how robots could support the process of inclusion among adults~\cite{strohkorb2020strategies}, and children~\cite{Gillet-RSS-20, tuncerSmileInclusion} or shape participation behavior \cite{charisi2021effects, Tennent2019, RobotLevelGilletCumbal2021}. 

These prior works show promise that robots can foster interaction among group members. However, prior literature also indicates that 
robot behaviours designed primarily based on existing (human-human interaction) literature might have different effects than originally hypothesized \cite{Martelaro2015, Strohkorb2016, short2017robot}. For example, a robot behaviour designed to reinforce group member performance differences rather than equalize them led to higher perceived cohesion - inverting the researchers' hypothesis \cite{short2017robot}. This work investigates whether we can leverage an extended co-design process with in-situ use and demonstration of robot behaviours to tackle this difficulty in developing effective robot behaviours for groups.
Prior literature that considers repeated interactions between robots and groups is sparse. However, Levinson et al. \cite{Levinson2020LearningRobots}, using a group-robot setup similar to ours, explored how children in groups of nine could improve language skills during a 3-week summer school. That work provides positive initial evidence of the benefits of a `group assistant' robot in this type of interaction setting, motivating the group assistant use case we chose to explore with our participants. 

\subsection{Participatory, End-to-End Robot Design (with Teenagers)}
Lee et al. provide an overview of how participatory design (PD) methods can be applied to social robot design, highlighting specifically how mutual learning, i.e., two-way exchange between researchers and end-users, can empower the latter to become active robot co-designers~\cite{leeStepsParticipatoryDesign2017}. This concept of mutual learning, in line with an overall mutual shaping approach~\cite{vsabanovic2010robots}, is fundamental to our \textit{summer school-research study} design. We aimed to learn from teenage participants while simultaneously educating them about social robotics and AI. This approach empowered participants then to lead our co-design process (the \textit{research study} part). Further, participants left the summer school with an understanding and experience of such that goes beyond `just' our research study (the \textit{summer school} part).  

Specifically implementing human-centred PD with teenagers, Björling et al. demonstrated how PD methods could be used to engage with teenagers meaningfully~\cite{bjorling2019participatory,bjorling2020exploring}. Especially pertinent to our current work, this included teens' in-situ wizarding of co-designed robots as part of the participatory design process. More broadly, the value of working to co-design robots \textit{for} children \textit{with} children has been repeatedly demonstrated in HRI (see e.g.~\cite{alves2021children}). However, to our knowledge, PD with young people is yet to be applied to the participatory design of robots for groups. 

Recent work in social HRI has specifically tried to incorporate PD, mutual learning, and in-situ evaluation processes into a single, generalisable, end-to-end design, automation and evaluation methodology. This methodology thereby aims to empower non-roboticist, domain experts to lead all stages of robot design and development (LEADOR: Led by Experts Design and Automation Of Robots~\cite{winkle2021leador}). LEADOR essentially has two key stages; the first stage is focused on using PD to identify what the robot should be able to do and what inputs the robot should use. Notably, this stage does \textbf{not} try to identify any specific rules connecting the two. This connection is instead addressed in the second stage. The ``naive'' robot  is deployed and ``taught" how to behave by a domain expert (e.g., using machine learning) during real interactions with end-users. 

For our chosen application of robots for teenage groups, recruiting, e.g. a schoolteacher or a child psychologist to take on the domain expert role would be more in line with how this methodology has been applied previously~\cite{winkle2020situ,senft2019teaching}. However, we decided to emulate the participatory design approaches taken by Björling et al. in their work with teenagers~\cite{bjorling2019participatory,bjorling2020exploring} further motivated by UNICEF's recent policy guidance highlighting the importance of including children and young people in the design of AI-powered systems. In summary, our long-term aim is to pursue a LEADOR-like process that centres on children, our domain experts, per those best practices from Björling et al. ~\cite{bjorling2019participatory,bjorling2020exploring}, with this work being the first step in that direction. 

%% file: sections/03_hypothesis.tex
\section{Methodology}
\label{sec:school}
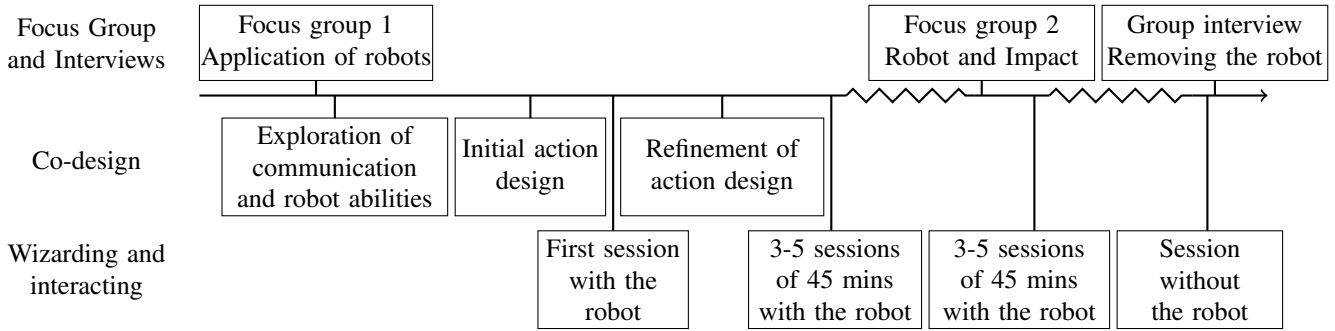
\begin{figure*}[h]
    \centering
    \input{figures/overview_methods}
    \caption{This timeline provides an overview of the different activities and methods used to explore a social robot group assistant for teenagers.}
    \label{fig:study_overview}
\end{figure*}

We designed our study activities 
to broadly address the following research question:
\vspace{0.4em}

    \textit{(How) do teens envisage a social robot improving their group experience, and (how) does this view evolve when controlling and interacting with such a robot across multiple group interaction sessions?}
\vspace{0.4em}

In order to address this research question, we combined typical PD focus group and interview sessions with robot group working/wizarding sessions per our setup in Figure~\ref{fig:training}. Figure \ref{fig:study_overview} details the progression of activities over the time of our study. Each activity is further detailed in an individual section below. 

Our overall approach was fundamentally centred around empowering students to lead in designing a peer-designed robot group assistant. To this end, we refrained from specifying what our proposed group assistant robot should be trying to achieve beyond making group working \textit{`better'}. We left it to participants to tell us what `good' versus `bad' group working looked and felt like, how group working could be made better and what role, if any, a social robot might play in achieving this. In this article, we focus predominantly on qualitative insights drawn from participants' experiences, as well as our own observations of group interactions. 



\subsection{Focus Group 1: Application of the robot}
The goal of the first focus group was to capture teenagers' expert knowledge with respect to groups prior to any interactions with the robot or presentation material from the researchers to explore initial ideas of how social robots could support and improve group work. The session was supported by an interim researcher-led presentation per~\cite{winkle2020mutual} introducing examples of robots/applications from related literature on robots for groups. Researchers introduced several robots as well as applications, i.e. \cite{Shen2018, Tennent2019, Strohkorb2016,StrohkorbSebo2018,Short2017Understanding}. For example, we introduced a version of Keepon designed to reduce object possession conflicts~\cite{Shen2018} and the Micbot, which was found to balance engagement~\cite{Tennent2019}.

\subsection{Exploration of verbal and non-verbal communication}
We engaged teenagers in a variety of activities and games, for example, an adapted version of the `telephone game'\footnote{\url{https://en.wikipedia.org/wiki/Chinese_whispers}} where participants had to pass a message with each person using a different modality (movement, facial expression, sound etc.). With these activities, we aimed to create awareness for different forms of verbal and non-verbal communication (both in human-human and human-robot interactions). We intended that this awareness would help participants to design the robot's multimodal actions.
\subsection{Initial action design and refinement}
The iterative action design process started with an initial brainstorming session within groups. These initial ideas were then presented and negotiated among all participants to decide upon the final set of abstract robot actions. After testing these actions in one group working session, we invited participants to refine the originally designed actions. Further, participants designed specific action instances, which they decided to pool across groups for the largest possible action variety. Examples of these are given in Table \ref{tab:actionspace}.
\subsection{Robot group working sessions}
The group working sessions (time-wise) made up the largest element of the study period. In-line with the LEADOR methodology, we briefed participants that these group working sessions were \textit{training sessions} for the robot; as we would eventually like to use machine learning to make the robot autonomous based on what they ``taught it". We explained why this meant we needed to undertake multiple and repeated sessions. The evaluation of such an autonomous robot based on the training data collected during this study is still future work. Therefore, we refer to these sessions instead as robot group working sessions.

To answer the research question set out in this paper, we observed how RC participants utilised the robot in these group working sessions and the impact on the group (equivalent to a multi-session HRI study). 

\begin{figure}[tbp]
\input{figures/tablet_replace}
\caption{Tablet interface participants' used to control the robot (utilising the co-designed action space) during group activity sessions. Users' had to select the desired action and could identify an individual target GM where appropriate. Selecting an action only (with no individual target) resulted in an instance of the group-targeting version of that action.}
\label{fig:tablet}
\end{figure}
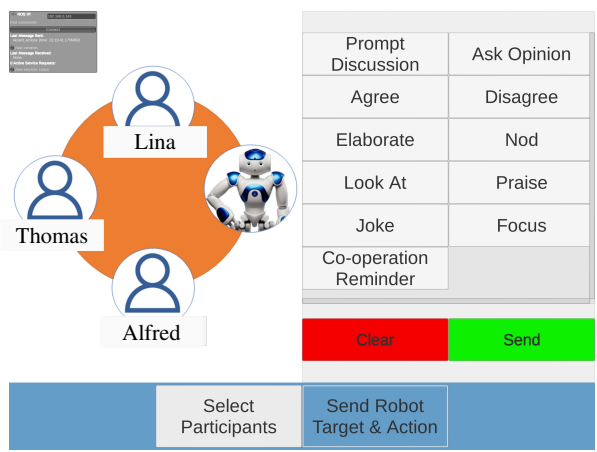

The robot controller (RC) role is equivalent to the \textit{domain expert teacher} referred to in the LEADOR method~\cite{winkle2021leador} as they ``teach" the robot what to do by using a tablet to ``tell" the robot when to do which action (and to whom address it). The tablet interface used to control the group assistant robot is depicted in Figure \ref{fig:tablet}. The GO role was designed to reflect the \textit{witness} role utilised by Björling and Rose~\cite{bjorlingExploringTeensRobot2020} when undertaking in-the-wild robot studies with teenagers. 
The setup was fixed as per Figure~\ref{fig:training}. Whilst 3 out of 5 group members were working on a discussion-based group activity, one group member was controlling the robot's actions. The final group member was observing the group to support subsequent reflections. Working with the robot was conducted in an adjoining, separate room to the main classroom. We conducted whole class activities and hosted complimenting summer school activities  in the main classroom for those not currently working with the robot.
The setup was fixed as per Figure~\ref{fig:training}. Whilst three of the five group members were working on a discussion-based group activity, one group member was controlling the robot's actions.  The final group member was observing the group to support subsequent reflections. These group robot sessions were conducted in an adjoining, separate room to the primary classroom. In the primary classroom, we conducted whole class activities and hosted complimenting summer school activities for those not currently working with the robot.

For each robot group working session, the students were left to work primarily unsupervised once the task had been explained. 
Researchers sat in the adjacent primary classroom and occasionally checked on the group via the door window without interrupting. Sessions comprised 3x15 minute rotations, in between which participants took quick breaks to complete short post-session evaluations and swap roles, rotating between being the RC, GO or a GM. The group activities used for these sessions included a number of in-house designed discussion-based activities\footnote{available at \url{https://kwinkle.github.io/resources.html}} plus adaptations of publicly available activities (such as the MIT AI ethics curriculum\footnote{An Ethics of Artificial Intelligence Curriculum for Middle School Students was created by Blakeley H. Payne with support from the MIT Media Lab Personal Robots Group, directed by Cynthia Breazeal}).

\subsection{Focus group 2 - Reflection on the Robot and its Impact}
This focus group was primarily focused on re-visiting the topics from Focus Group 1. Since participants had the chance to experience both controlling and interacting with the robot, we aimed to reflect on their unique insights following this experience, i.e. what (if any) impact the robot had during group working and whether this matched their initial expectations. 
\subsection{Group interview after removing the robot}
We wanted to explore the extent to which, if at all, participants found the robot to be useful in supporting their group activities. Given the large number of robot group working sessions participants engaged in, we asked participants to engage in one final group working session \textit{without} the robot. Afterwards, we invited participants to reflect upon this experience. 
This experience made for an interesting comparison to discussions from Focus Group 2, which was also conducted after experiencing a number of group working sessions but before this `no robot' session. 

\subsection{Approach}
We set out to implement the recommendations put forward by Björling et al. \cite{bjorling2019participatory} and follow UNICEF guidance on AI for children. These recommendations shaped how we briefed participants about the study and our experimental setup. We were completely upfront with participants about the nature of our research study at all times. This approach applied both to how we presented the purpose the study was embedded in (i.e., our intent to `do research') and the transparency of our experimental setup. The use of cameras and microphones was communicated to (and purposefully \textit{not} hidden away from) participants. Further, particularly emulating~\cite{bjorlingExploringTeensRobot2020}, we did not attempt to hide or even distance the RC from the rest of the group during the group interaction sessions (see Figure~\ref{fig:training}). 

\subsection{Data coding and extraction}
Focus group and interview sessions were automatically transcribed and corrected by the researchers. Resultant data were coded for key results following the Framework Method~\cite{ritchie2013qualitative} i.e., using a combined deductive and inductive approach to coding. We, thereby, were primarily guided by the inductive approach using only a couple of high-level initial codes generated in line with the research question. Three of the authors then independently coded the data collected from the participant group they had worked with most. Data-driven inductive codes were generated as required. The results were discussed, and a final coding scheme was generated for application to all data.

The data used to analyze the use of the different actions was collected directly from the system. The system automatically noted which action was executed from the tablet and to whom the action was addressed. Further, we noted who was RC in each of the sessions.

\subsection{Participants and Groups}
We advertised the summer school research study via social media, through our internal university communication channels and through our research network of teaching staff at local schools. Places were primarily allocated on a first come first served basis whilst we also attempted to foster some diversity with regards to participant age, gender and school, thus allowing creation of more heterogeneous groups. We received a total of 24 applications (2 of which were ineligible based on applicant age) and ultimately recruited 16 participants (8 boys, 8 girls, ages 12-15 with M = 12.8) to take part over the two weeks, of whom 14 participants attended all 10 days without absence. Consent forms were collected from each participant in addition to a parent/guardian. 

The study was conducted in Sweden. However, the language of communication was English which applicants were informed about in the advertisement material. Whilst we are cognizant that this may have been a barrier to participants who do not feel comfortable speaking English (although Sweden ranks very highly for English proficiency\footnote{ranking 8th out of 112 countries in 2021: https://www.ef.se/epi/}) it removed the (more common) Swedish language barrier for participants who e.g. come from newly arrived families and/or attend international, English speaking schools.


Participants did not pay/were not paid to attend the summer school, but an on-campus lunch was provided every day. Moreover, participants were gifted a university `goodie bag' with pen, mug etc. on completion of the summer school. 

Participants were randomly assigned to one of three groups (\textit{Wall-E}, \textit{R2D2}, and \textit{Baymax}) primarily under the supervision (for researcher-led activities/practical management purposes), of the first, second and third authors, respectively. Hereafter, we will refer to each participant with a unique ID whose first letter represents the group they belong to. 

%% file: figures/overview_methods.tex
\begin{tikzpicture}
   \node[text width=2.3cm, align=center] at (-1.5, 0.7)  (a2) {Focus Group and Interviews};
   \draw[thick] (1.55,0.2) -- (1.55, 0);
   \draw  (0,0.2) rectangle (3.1, 1.2) node[ pos=.5,text width=4cm,align=center] {Focus group 1 \linebreak Application of robots};
   \draw[thick] (10.4,0.2) -- (10.4, 0);
  \draw  (8.9,0.2) rectangle (11.9, 1.2) node[ pos=.5,text width=4cm,align=center] {Focus group 2\linebreak Robot and Impact};
 
 \draw  (12,0.2) rectangle (15,1.2) node[ pos=.5,text width=4cm,align=center] {Group interview\linebreak Removing the robot};
 \draw[thick] (13.5,0.2) -- (13.5, 0);

    \node[text width=2.3cm, align=center] at (-1.5, -0.9)  (a2) {Co-design};
   \draw[thick] (1.8,-0.3) -- (1.8, 0);
   \draw  (0.3,-0.3) rectangle (3.3, -1.6) node[ pos=.5,text width=4cm,align=center] {Exploration of communication and robot abilities};
   \draw[thick] (4.4,-0.3) -- (4.4, 0);
  \draw  (3.4,-0.3) rectangle (5.4, -1.6) node[ pos=.5,text width=2.5cm,align=center] {Initial action design};
  \draw[thick] (6.95,-0.3) -- (6.95, 0);
 \draw  (5.6,-0.3) rectangle (8.3,-1.6) node[ pos=.5,text width=4cm,align=center] {Refinement of action design};  
  
       \node[text width=2.3cm, align=center] at (-1.5, -2.35)  (a2) {Wizarding and interacting};
   \draw[thick] (5.5,-1.8) -- (5.5, 0);
   \draw  (4.5,-1.8) rectangle (6.5, -3.1) node[ pos=.5,text width=2cm,align=center] {First session with the robot};
  
  \draw[thick] (8.4,-1.8) -- (8.4, 0);
 \draw  (7.3,-1.8) rectangle (9.6,-3.1) node[ pos=.5,text width=2.2cm,align=center] {3-5 sessions of 45 mins with the robot
};
 \draw[thick] (11.1,-1.8) -- (11.1, 0);
  \draw  (9.7,-1.8) rectangle (12.1, -3.1) node[ pos=.5,text width=2.2cm,align=center] {3-5 sessions of 45 mins with the robot
};
\draw[thick] (13.4,-1.8) -- (13.4, 0);
 \draw  (12.2,-1.8) rectangle (14.4,-3.1) node[ pos=.5,text width=2.2cm,align=center] {Session without the robot};

  \draw[thick] (0,0) -- (8.6, 0);
  \draw[snake, thick] (8.6,0) -- (10.3,0);
  \draw[thick] (10.3,0) -- (11.3, 0);
  \draw[snake, thick] (11.3,0) -- (13.2,0);
  \draw[thick,->] (13.2,0) -- (14.2, 0);
\end{tikzpicture}

%% file: figures/tablet_replace.tex
\begin{tikzpicture}
        \path (0,0) node(a) {\centerline{\includegraphics[width=0.9\linewidth]{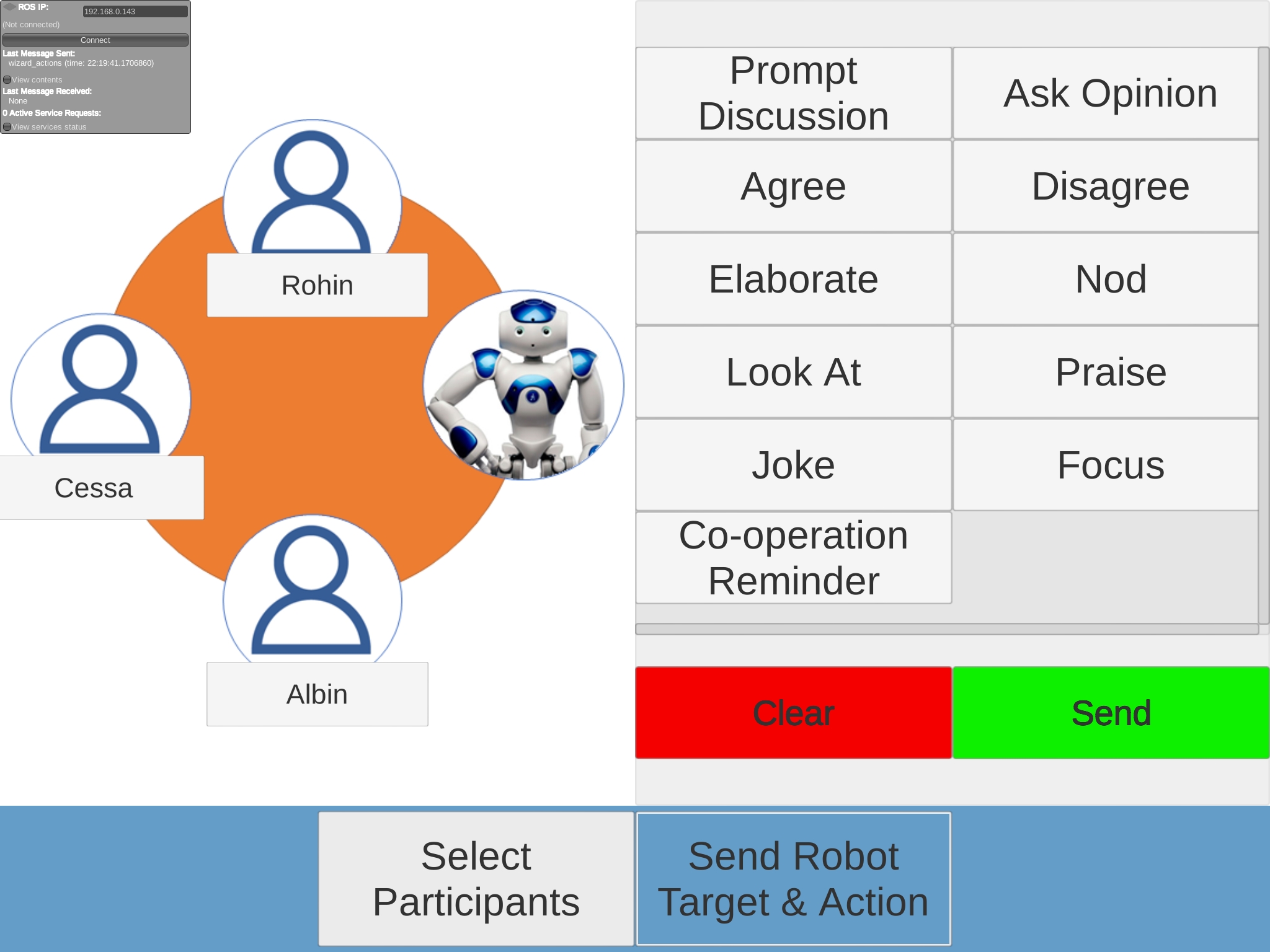}}};
        \node[rectangle, draw,  white!98!black, fill = white!98!black, text=black, minimum width=1.6cm, scale=0.85] (r) at (-2,-1.35) {Alfred};
        \node[rectangle, draw,  white!98!black, fill = white!98!black, text=black, minimum width=1.6cm, scale=0.85] (r) at (-1.95,1.18) {Lina};
        \node[rectangle, draw,  white!98!black, fill = white!98!black, text=black, minimum width=1.6cm, scale=0.85] (r) at (-3.32,-0.06) {Thomas};
\end{tikzpicture}  

%% file: sections/04_envisioning_robot_use.tex
\section{RESULTS} \label{sec:focus}
The longitudinal nature of this study allowed us to capture initial impressions as well as if/how the impact and usefulness of the robot changed after interacting with it in several robot group working sessions. We first explored how teens envisage a robot group assistant and possible applications before they had experienced working with the robot. We then used Focus Group 2 and the Group Interview to see whether the robot ``lived up to expectations", and if/how this changed over time and/or between groups.

\subsection{Applications for a Robot Group Assistant}\label{sec:focus_findings_use}
Ideas on how a robot might help to improve their groups broadly fell into three categories: \textit{prompting discussion}, \textit{turn taking} and \textit{improving efficiency/performance}. 
\begin{description}[align=left,leftmargin=0em,labelsep=0.2em,font=\textbf,itemsep=0em,parsep=0.3em]

 \item[Prompting discussion]: All groups thought that a
 a robot could 
 act as an ice-breaker and to ease awkwardness: \textit{``It's mostly about starting that conversation and then it'll take on and then everyone else will start talking"} (W2).
\textit{``And they could be like how you do it, or in math. It's like it's a group. They're like, OK, let's get started like they talk for us.''} (W4)

\item[Turn taking]: Participants also discussed that the robot could help so that everyone got to speak and perhaps encourage those who were quiet: \textit{``If someone talks way too much it's like: sh sh sh, let this one talk"} (W3).
This idea of a robot for turn-taking and, more specifically, to improve inclusiveness was strengthened after the researcher-led presentation.
\textit{Micbot}, a robot microphone which rotates in order to try and equalise speaking time and participation during group discussions~\cite{Tennent2019}, was particularly well-received (a favourite within R2D2 and Baymax). Participants were impressed at how simple yet effective it was at getting all group members to participate in the discussion: \textit{``Well, what I liked about the Micbot...it gave everyone a voice. In a way that I think that I don't really feel like [Keepon \cite{Shen2018}] could do. I don't know how they could do it in the same way"} (R1). This potential for the management of turn-taking to increase inclusiveness seemed to appeal equally to participants who self-identified as (and were observed to be) both more extroverted/confident and more introverted/shy: \textit{``And it's going to let the shy person be able to talk...and everyone after might not speak over him"} (W4, more extroverted); \textit{``I can't interrupt them and then I am screwed. So I wait and wait and wait. So this microphone will be really good so I could talk to the others"} (W2, more introverted).

\item[Improving efficiency/performance]: A common theme centred on how the robot could help with the task by answering questions or acting as an interface to the internet: \textit{``Like a computer, it tells you the answers. You have to work on an assignment you have the robot that is like the computer or google, it can help you to find out the answer for your assignment''} (B4). 
\end{description}

\subsection{Design and Use of the Robot's Action Space}

The three key functions \textit{prompting discussion}, \textit{turn taking} and \textit{improving efficiency/performance} described in \ref{sec:focus_findings_use} 
are clearly reflected in the actions participants designed and refined for NAO during our iterative action design process (Table~\ref{tab:actionspace}).
\input{tables/table_codesigned_actionspace}
It was infeasible to implement, e.g. real-time internet search via the robot. However, the use of the robot for improving efficiency and performance seemingly still emerged in the form of actions meant to keep the group `on task'. Participants created different actions centred around prompting the group (or specific individuals within the group) to stay focused in working towards task completion (see the Focus action in Table~\ref{tab:actionspace}). 
After some initial testing, participants quickly suggested that the robot should always refer to an individual by name when delivering an individual-targeted action in addition to looking at them.
This design choice had not initially been specified in any of the action instances they had created, seemingly becoming clear only upon working with the robot directly. We therefore appended target name in every (speech-based) individual-targeting action instance (see examples in Table~\ref{tab:actionspace}) during the action refinement phase. 

\begin{figure}[t]
\centerline{\includegraphics[height=4.2cm]{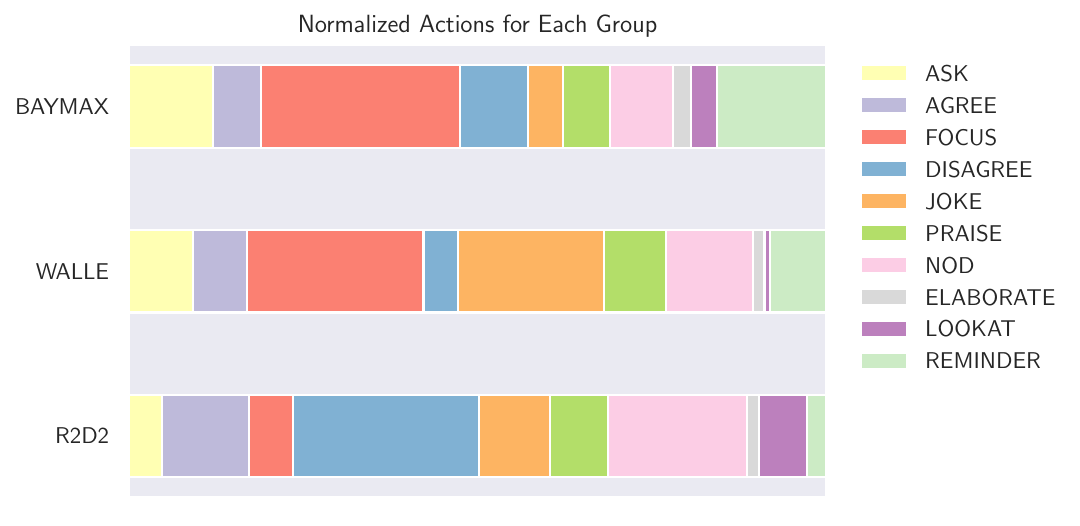}}
\caption{Use of the robot's actions across the groups. The use of actions reflects observed differences in how those groups behaved together and how therefore the robot could help best. 
Best viewed in color.}
\label{fig:actionsgroups}
\end{figure}

The way that different RCs utilised this common action space during group working sessions varied quite significantly between groups, reflecting that the groups were quite different and distinct in their interaction. 
For instance, as can be seen in Fig.~\ref{fig:actionsgroups}, RCs in the Baymax group used the robot in a teacher-like manner: prompting team members to focus on the required task and nudging them toward cooperation. 
The R2D2 group, who appeared (and self-identified as being) disciplined and capable of self-managing their teamwork, seemingly used the robot instead in a peer-like manner, using it more like a source of fun and entertainment. 

\begin{figure}
    \centering
    \includegraphics[height=4.2cm]{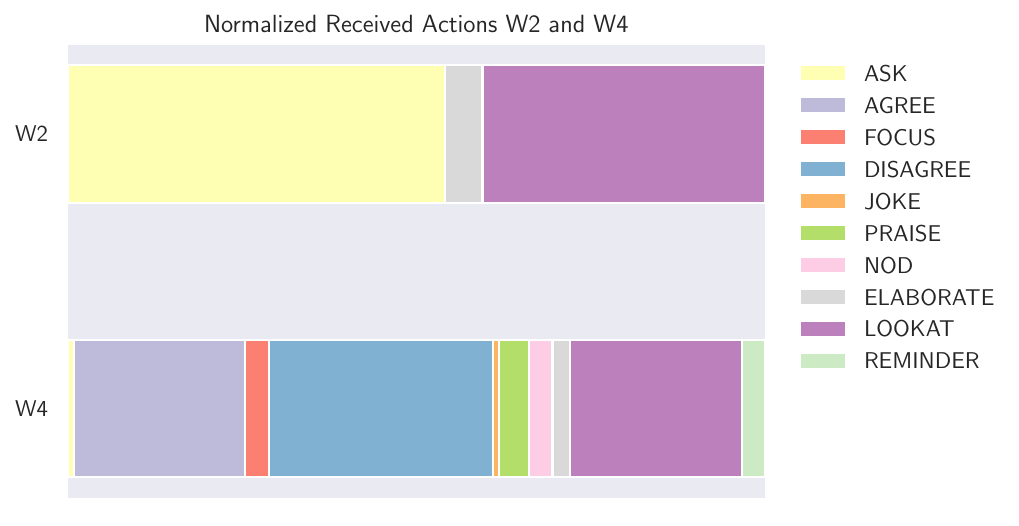}
    \caption{Use of the robot's action space toward two very different participants (extroverted W4 versus introverted W2). The received actions seemingly reflect their different personalities. Best viewed in color.}
    \label{fig:sfig1}
\end{figure}

As well as responding to the current interaction in the group, RCs' action choices reflect their personalities and those of GM(s) their action choices target. For example, Figure~\ref{fig:sfig1} shows the actions 
targeted towards two participants within the Wall-E group. 
The actions addressed towards W2 (self-identified introvert, often observed struggling to take part in the conversation) by the RCs seemingly tried to increase her engagement as they asked for her opinion or looked to her to elaborate more on ideas/opinions.

In contrast, W4 (self-identified extrovert, observed to be more often leading group discussions) received a wider variety of actions. Especially, W4 received those that might be considered indicative of greater (unprompted) participation but also perhaps an inclination to engage in off-topic discussions or become distracted/over-excited (i.e. \textit{agree, disagree, focus} and \textit{praise}).  


\subsection{Expectations vs. Reality}
After working with the robot over the multiple group robot sessions, participants were able to reflect on its usefulness critically. Interestingly, each group seemingly experienced this usefulness differently throughout the study, so we present the following findings by group. 

In the initial focus group, R2D2 demonstrated some hesitation as to whether a robot was necessary/helpful in groups that already know each other: \textit{``When you don't know anyone that I think maybe robots can be like good at like organising and making sure that everyone can talk... I feel like if you know the people you are in a group with it I feel like it would more be like the group and the robot, not like the robot helping the group"} (R2).

In their reflections on the robot's impact, they generally described being impressed by what the robot could do to improve group work. Moreover, they confirmed their initial doubts about the robot's utility once the group got to know each other better and worked quite well together. \textit{``Mine are much more positive for robots. I think that they can actually help to like recover a bad discussion and also I feel like it has more potential than I thought in the beginning''} (R2) \textit{``I feel like it's gotten less helpful. Just because we got to know each other"} (R4) \textit{``Yeah, maybe sometimes it still helps. When it's like asking someone for an opinion, I think but it's much more rare these days''} (R2).

After working without the robot, R2D2 still thought that they did not \textit{need} the robot but noted it improved their enjoyment: \textit{``These last days it has like worked well even like the robot hasn't had the need to do much. So it wasn't like a worse discussion, but it's more fun when the robot is there. So you get like more energy."} (R2) \textit{``Exactly"} (R1).
\input{tables/removal_comments}
Baymax shared similar reflections to R2D2 when discussing the robot's changing impact on their interactions. They highlighted a change in respect to group members becoming less hesitant to speak during group activities and hence less in need of prompting. However, they also identified ways in which 
the robot did still actively help, specifically in the management of turn-taking and keeping the group focused. \textit{``I think that it still is helpful because it can still help with who's going to talk now. Maybe this person has been a little bit talking. This person has been quiet. It gets focused, get back to the topic. I think it's still really good like that, but I think that it's not that much necessary anymore, because now everybody is talking"} (B2) \textit{``Yeah, I also think that the robot it’s pretty helpful. Because if the group is not focusing then it can actually help the group to focus''} (B3) \textit{``It's helpful, but not as much anymore''} (B5).

These ideas were (at least somewhat) challenged when the group worked together without the robot, as reflected in their comments in Table \ref{tab:norobotsessions}. All Baymax members seemed to agree that they did not work well together in this session and reflected that actually, the robot was important in helping them to work together well: \textit{``The robot, always reminds you focus, do things, ask questions, but now they just like, it's so easy to just don't work."} (B4) \textit{``I think that it was really like a lot of people just lost focus and started talking about something else. I was the only one working, actually."} (B2)

Finally, Wall-E had mixed opinions. Seemingly the impact of the robot varied depending on which three participants took the GM roles. For example (W4, observed to very much engage with the task) lamented the robot's absence after conducting group work with two less task-engaged group members: \textit{But you know who you are, they kept on arguing, right? Yeah, and the robot could have stopped that, yeah, so I had to... The robot could have said like focus up. Or something to make them stop, but there was no robot so I had to try to stop them and it didn't work so.''} (W4) \textit{``I mean, it depends, we are not always in the same group. Different groups work different"} (W1). One Wall-E member specifically raised an interesting idea regarding the robot actually being \textit{more} important for groups who know each other rather than for groups working together for the first time, which was at odds with their previous ideas (as well as R2D2s comments) on where such a robot might be most helpful: \textit{``If you know people very well, I think it would help like more than if you do not know them very well... because in first time interactions you wouldn't be like screaming at them, you would be a little calm to get to know them.''} (W5)

%% file: tables/table_codesigned_actionspace.tex
\begin{table}[]
\caption{Final co-designed action space of the group assistant robot. Actions default targeted the entire group unless the robot controller indicated a specific individual target via the teaching tablet (Figure \ref{fig:tablet}).}
\begin{center}
\begin{tabular}{|p{1cm}|p{1.2cm}|p{5cm}|}
\hline
Action                     & Target     & Example Speech (written by participants) \\ \hline
{Ask Opinion}       & Group      & Let's discuss our opinion about this! \\ \cline{2-3} 
                          & Ind. & Do you have an opinion on this matter [name]? \\ \hline
{Agree}     & Group      & I agree.  \\ \cline{2-3} 
                          & Ind. & I agree with you [name] \\ \hline
{Focus}     & Group      & Please can we stay focused on the activity? \\ \cline{2-3} 
                          & Ind. & Don't forget the task [name]! \\ \hline
{Disagree}  & Group      & I don't agree with this. \\ \cline{2-3} 
                          & Ind. & I don't agree with you [name] \\ \hline
{Joke}      & Group/Ind  & I wrote a song about a tortilla. Actually, it's more of a wrap! \\ \hline
{Praise}    & Group      & Great job everyone! \\ \cline{2-3} 
                          & Ind. & "Wow! Good thing you brought that up [name]!" \\ \hline
{Nod}       & Group/Ind    & - \\ \hline
{Elaborate} & Group      &  Really? Why do you think so? \\ \cline{2-3}
                          & Ind. & Can you maybe explain more what you mean [name]? \\ \hline
{Look At}   & Group/Ind.      &  - \\ \hline
{Reminder}  & Group      & Remember: teamwork divides the task and multiplies the success. \\ \cline{2-3} 
                          & Ind. & [Name] remember that we are a team, we have to cooperate. \\ \hline
\end{tabular}
\end{center}
\label{tab:actionspace}
\end{table}

%% file: tables/removal_comments.tex
\begin{table*}[]
\caption{A selection of (anonymous) post-session group member and observer evaluation scores and comments from the group activities undertaken (without the robot being present) on Day 9. Observer evaluations did not have a question on activity enjoyment. E = emoji-based Likert (reported here as 1-5 with 5 being best) and 100 = scalar measures out of 100.}
\resizebox{\textwidth}{!}{
\begin{tabular}{|l|l|l|l|l|}
\hline
Enjoyment (E) & Work (E) & Robot Help (100) & Want Robot (100) & Comments                                                                                                                                                                                                                                                                                 \\ \hline
\multicolumn{5}{|l|}{\textbf{Baymax}}                                                                                                                                                                                                                                                                                                                            \\ \hline
5         & 1               & 100              & 100                & I was the only one working we need the robot.                                                                                                                                                                                                                                            \\ \hline
2         & 1               & 100              & 100                & It could make me concentrate again.                                                                                                                                                                                                                                                      \\ \hline
1         & 3               & 100              & 100                & \begin{tabular}[c]{@{}l@{}} I think the robot could help us to work harder. We didn't focus that much without \\ the robot. \end{tabular}                                                                                                                                                                                              \\ \hline
-         & 2               & 100              & 100                & \begin{tabular}[c]{@{}l@{}}They didn't focus. It could remind them to stay focused and co-operate. I think \\ they would prefer it because they seemed pretty uncomfortable.\end{tabular}                                                                                                 \\ \hline
\multicolumn{5}{|l|}{\textbf{R2-D2}}                                                                                                                                                                                                                                                                                                                           \\ \hline
4         & 1               & 10               & 25                 & We didn't really need any help. It is just more fun with the robot in the room :)                                                                                                                                                                                                        \\ \hline
3         & 3               & 100              & 100                & It was boring without NAO. It's just better {[}when it's here{]}.                                                                                                                                                                                                                        \\ \hline
-         & 4               & 20               & 30                 & It worked so well on it's own but the focus button could have helped.                                                                                                                                                                                                                    \\ \hline
-         & 5               & 20               & 30                 & The argued a bit and it could have helped them not to.                                                                                                                                                                                                                                   \\ \hline
\multicolumn{5}{|l|}{\textbf{Wall-E}}                                                                                                                                                                                                                                                                                                                          \\ \hline
-         & 1               & 100              & 100                & They did not even work they just talked. Without the robot they didn't work.                                                                                                                                                                                                             \\ \hline
4         & 4               & 83               & 93                 & \begin{tabular}[c]{@{}l@{}}The group worked well however some people were arguing. The robot would stop \\ the arguing.\end{tabular}                                                                                                                                                      \\ \hline
5         & 5               & 0                & 20                 & \begin{tabular}[c]{@{}l@{}}We all agreed on most stuff and had fun and were focused. The robot keeps \\ disagreeing with people which makes everyone unfocused. If there wasn't a robot \\ controller maybe the robot would help. We didn't need it but it could be more fun.\end{tabular} \\ \hline
3         & 3               & 0                & 100                & We would have argued anyway but it could have made my opinions stronger.                                                                                                                                                                                                                 \\ \hline
\end{tabular}}
\label{tab:norobotsessions}
\end{table*}

%% file: sections/06_discussion.tex
\section{Design Implications: Robots as Ice-Breakers, Turn-Takers and Fun-Makers}
Many of our results on using robots to improve teen groups revolve around turn-taking, participation and inclusion. Theories on group development describe how each group and its processes move through stages (forming, storming, norming, performing, adjourning)~\cite{tuckman1965developmental}. Interestingly, this theory is reflected in the teen's views and designs of social robot behaviours. For new groups which are still in the stage of forming, the robot would represent an ice-breaker, something to help reduce the awkwardness by asking questions and indicating who should speak. Whilst there was some hesitation over whether this would still be useful for established groups who knew each other, the robot group working sessions demonstrated that (for two of our three groups) this turn-taking (combined with, e.g. focus reminders) became crucial for including all group members in a discussion (and keeping them on task). As these groups got past that initial awkwardness and seemingly moved through the stages of storming and norming, 
the \textit{politeness} driven turn-taking perhaps started to decrease, so we saw the role of the robot evolve from an ice-breaker eliciting discussion to more of a turn manager, ensuring all participants (those more shy and those easily distracted) 
contributed to discussions. 

However, we also observed differences in how the specific groups behaved throughout the stages, seemingly resulting in different robot behaviours. R2D2 appeared to very well self-manage turn-taking and focus. For this group, the robot appeared to evolve into a fun-maker, being used to inject fun into the group work. 
Based on these observations, we conceptualise our groups as being along a spectrum regarding their need/use of the robot. Where for R2D2 the robot was not really \textit{needed} but still somehow added to the experience, for Wall-E the use/necessity of the robot varied based on the specific sub-combination of group members. Consequently, for Baymax, the robot played a key role in helping the group work together (Baymax). 

We suggest it might be useful to consider this spectrum when designing for group-robot interactions. Future work should explore how specific group dynamics (e.g. different group member personalities), time spent working with the robot/as a group, and other dynamic factors might influence where a group is likely to be on this spectrum. These observations pose opportunities and challenges for creating robots that can move dynamically along the spectrum, taking on an ice-breaking, turn-taking, or fun-making role, both between and within groups at different stages of their development.

Detailed reflection on whether our setup could facilitate autonomy via machine learning is out of scope for this article. However, our participants' use of the jointly designed action space (and the post-session evaluation forms) demonstrates that they were generally thoughtful RCs. They seemingly tailored their use of the robot's actions based on their knowledge of their group members and the dynamics they were observing. Further work is required to fully examine the feasibility of creating robust and effective autonomous social robot behaviour from this data. 

%% file: sections/07_conclusion.tex
\section{Conclusion}
This article presents insights from a study centred on the design and (longitudinal) evaluation of social robot group assistants. Our only design brief to the participants was that the robot should simply make group working `better', working with participants to decide exactly what that meant. 
Whilst our study was still relatively short-term, each group spent $\sim$10 hours working with the robot. Our results provide insight beyond the initial novelty factor and demonstrate that even if the role of the robot evolves along this time dimension, it does not necessarily become obsolete for those groups that do not `need' it. Instead, groups may posit (and dynamically move) along a spectrum regarding their needs/wants from a group assistant robot. 
Future work will explore how these design insights and the extensive training data collected, can create meaningful, peer-designed robots that can assist as ice-breakers, turn-takers and fun-makers.

%% file: root.bbl
\begin{thebibliography}{10}
\providecommand{\url}[1]{#1}
\csname url@samestyle\endcsname
\providecommand{\newblock}{\relax}
\providecommand{\bibinfo}[2]{#2}
\providecommand{\BIBentrySTDinterwordspacing}{\spaceskip=0pt\relax}
\providecommand{\BIBentryALTinterwordstretchfactor}{4}
\providecommand{\BIBentryALTinterwordspacing}{\spaceskip=\fontdimen2\font plus
\BIBentryALTinterwordstretchfactor\fontdimen3\font minus
  \fontdimen4\font\relax}
\providecommand{\BIBforeignlanguage}[2]{{%
\expandafter\ifx\csname l@#1\endcsname\relax
\typeout{** WARNING: IEEEtran.bst: No hyphenation pattern has been}%
\typeout{** loaded for the language `#1'. Using the pattern for}%
\typeout{** the default language instead.}%
\else
\language=\csname l@#1\endcsname
\fi
#2}}
\providecommand{\BIBdecl}{\relax}
\BIBdecl

\bibitem{brown1987peer}
B.~B. Brown and M.~J. Lohr, ``Peer-group affiliation and adolescent
  self-esteem: an integration of ego-identity and symbolic-interaction
  theories.'' \emph{Journal of personality and social psychology}, vol.~52,
  no.~1, p.~47, 1987.

\bibitem{bosSocialNetworkCohesion2018}
W.~van~den Bos, E.~A. Crone, R.~Meuwese, and B.~G{\"u}ro{\u g}lu, ``Social
  network cohesion in school classes promotes prosocial behavior,'' \emph{PLOS
  ONE}, vol.~13, no.~4, p. e0194656, Apr. 2018.

\bibitem{Sebo2020RobotsReview}
S.~Sebo, B.~Stoll, B.~Scassellati, and M.~F. Jung, ``Robots in groups and
  teams: a literature review,'' \emph{Proceedings of the ACM on Human-Computer
  Interaction}, vol.~4, no. CSCW2, pp. 1--36, 2020.

\bibitem{Martelaro2015}
M.~F. Jung, N.~Martelaro, and P.~J. Hinds, ``{Using Robots to Moderate Team
  Conflict: The Case of Repairing Violations},'' in \emph{Proceedings of the
  Tenth Annual ACM/IEEE International Conference on Human-Robot
  Interaction}.\hskip 1em plus 0.5em minus 0.4em\relax Portland, Oregon, USA:
  Association for Computing Machinery, 2015, p. 229–236.

\bibitem{Tennent2019}
H.~Tennent, S.~Shen, and M.~Jung, ``{Micbot: A Peripheral Robotic Object to
  Shape Conversational Dynamics and Team Performance},'' \emph{ACM/IEEE
  International Conference on Human-Robot Interaction}, vol. 2019-March, pp.
  133--142, 2019.

\bibitem{RobotLevelGilletCumbal2021}
S.~Gillet, R.~Cumbal, A.~Pereira, J.~Lopes, O.~Engwall, and I.~Leite, ``{Robot
  Gaze Can Mediate Participation Imbalance in Groups with Different Skill
  Levels},'' in \emph{Proceedings of the 2021 ACM/IEEE International Conference
  on Human-Robot Interaction}.\hskip 1em plus 0.5em minus 0.4em\relax New York,
  NY, USA: ACM, 3 2021.

\bibitem{Strohkorb2016}
S.~Strohkorb, E.~Fukuto, N.~Warren, C.~Taylor, B.~Berry, and B.~Scassellati,
  ``{Improving human-human collaboration between children with a social
  robot},'' \emph{25th IEEE International Symposium on Robot and Human
  Interactive Communication, RO-MAN 2016}, pp. 551--556.

\bibitem{Short2017Understanding}
E.~S. Short, K.~Swift-Spong, H.~Shim, K.~M. Wisniewski, D.~K. Zak, S.~Wu,
  E.~Zelinski, and M.~J. Matari{\'c}, ``Understanding social interactions with
  socially assistive robotics in intergenerational family groups,'' in
  \emph{2017 26th IEEE International Symposium on Robot and Human Interactive
  Communication (RO-MAN)}.\hskip 1em plus 0.5em minus 0.4em\relax IEEE, pp.
  236--241.

\bibitem{leeStepsParticipatoryDesign2017}
H.~R. Lee, S.~Sabanovic, W.-L. Chang, S.~Nagata, J.~Piatt, C.~Bennett, and
  D.~Hakken, ``Steps {{Toward Participatory Design}} of {{Social Robots}}:
  {{Mutual Learning}} with {{Older Adults}} with {{Depression}},'' in
  \emph{Proceedings of the 2017 {{ACM}}/{{IEEE International Conference}} on
  {{Human-Robot Interaction}}}, ser. {{HRI}} '17.\hskip 1em plus 0.5em minus
  0.4em\relax {New York, NY, USA}: {ACM}, pp. 244--253.

\bibitem{alves-oliveiraYOLORobotCreativity2017}
P.~{Alves-Oliveira}, P.~Arriaga, A.~Paiva, and G.~Hoffman, ``{{YOLO}}, a
  {{Robot}} for {{Creativity}}: {{A Co-Design Study}} with {{Children}},'' in
  \emph{Proceedings of the 2017 {{Conference}} on {{Interaction Design}} and
  {{Children}}}.\hskip 1em plus 0.5em minus 0.4em\relax {New York, NY, USA}:
  {ACM}, Jun. 2017, pp. 423--429.

\bibitem{winkle2021leador}
K.~Winkle, E.~Senft, and S.~Lemaignan, ``{{LEADOR}}: {{A}} method for
  end-to-end participatory design of autonomous social robots,''
  \emph{Frontiers in Robotics and AI}, vol.~8, p. 343, 2021.

\bibitem{bjorling2019participatory}
E.~A. Bj{\"o}rling and E.~Rose, ``Participatory research principles in
  human-centered design: engaging teens in the co-design of a social robot,''
  \emph{Multimodal Technologies and Interaction}, vol.~3, 2019.

\bibitem{bjorling2020exploring}
E.~A. Bj{\"o}rling, K.~Thomas, E.~J. Rose, and M.~Cakmak, ``Exploring teens as
  robot operators, users and witnesses in the wild,'' \emph{Frontiers in
  Robotics and AI}, vol.~7, p.~5, 2020.

\bibitem{NonDyadicSchneiders22}
E.~Schneiders, E.~Cheon, J.~Kjeldskov, M.~Rehm, and M.~B. Skov, ``Non-dyadic
  interaction: A literature review of 15 years of human-robot interaction
  conference publications,'' \emph{J. Hum.-Robot Interact.}, vol.~11, no.~2,
  2022.

\bibitem{Shen2018}
S.~Shen, P.~Slovak, and M.~F. Jung, ``{"Stop. I See a Conflict Happening."},''
  in \emph{Proceedings of the 2018 ACM/IEEE International Conference on
  Human-Robot Interaction}.\hskip 1em plus 0.5em minus 0.4em\relax New York,
  NY, USA: ACM, 2 2018, pp. 69--77.

\bibitem{erel2021enhancing}
H.~Erel, D.~Trayman, C.~Levy, A.~Manor, M.~Mikulincer, and O.~Zuckerman,
  ``Enhancing emotional support: The effect of a robotic object on human--human
  support quality,'' \emph{International Journal of Social Robotics}, pp.
  1--20, 2021.

\bibitem{StrohkorbSebo2018}
S.~Strohkorb~Sebo, M.~Traeger, M.~F. Jung, and B.~Scassellati, ``{The Ripple
  Effects of Vulnerability: The Effects of a Robot's Vulnerable Behavior on
  Trust in Human-Robot Teams},'' \emph{Proceedings of the 2018 ACM/IEEE
  International Conference on Human-Robot Interaction - HRI '18}, no. February,
  pp. 178--186.

\bibitem{strohkorb2020strategies}
S.~Strohkorb~Sebo, L.~L. Dong, N.~Chang, and B.~Scassellati, ``Strategies for
  the inclusion of human members within human-robot teams,'' in
  \emph{Proceedings of the 2020 ACM/IEEE International Conference on
  Human-Robot Interaction}, 2020, pp. 309--317.

\bibitem{Gillet-RSS-20}
S.~Gillet, W.~van~den Bos, and I.~Leite, ``{A social robot mediator to foster
  collaboration and inclusion among children},'' in \emph{Proceedings of
  Robotics: Science and Systems}, Corvalis, Oregon, USA, July 2020.

\bibitem{tuncerSmileInclusion}
S.~Tuncer, S.~Gillet, and I.~Leite, ``Robot-mediated inclusive processes in
  groups of children: From gaze aversion to mutual smiling gaze,''
  \emph{Frontiers in Robotics and AI}, vol.~9, 2022.

\bibitem{charisi2021effects}
V.~Charisi, L.~Merino, M.~Escobar, F.~Caballero, R.~Gomez, and E.~G{\'o}mez,
  ``The effects of robot cognitive reliability and social positioning on
  child-robot team dynamics,'' in \emph{2021 IEEE International Conference on
  Robotics and Automation (ICRA)}.\hskip 1em plus 0.5em minus 0.4em\relax IEEE,
  2021.

\bibitem{short2017robot}
E.~Short and M.~J. Mataric, ``Robot moderation of a collaborative game: Towards
  socially assistive robotics in group interactions,'' in \emph{2017 26th IEEE
  International Symposium on Robot and Human Interactive Communication
  (RO-MAN)}.\hskip 1em plus 0.5em minus 0.4em\relax IEEE, 2017, pp. 385--390.

\bibitem{Levinson2020LearningRobots}
L.~Levinson, O.~Gvirsman, I.~M. Gorodesky, A.~Perez, E.~Gonen, and G.~Gordon,
  ``{Learning in Summer Camp with Social Robots: A Morphological Study:
  Studying Dynamics Using Social Robots},'' \emph{International Journal of
  Social Robotics}, 2020.

\bibitem{vsabanovic2010robots}
S.~{\v{S}}abanovi{\'c}, ``Robots in society, society in robots,''
  \emph{International Journal of Social Robotics}, vol.~2, no.~4, pp. 439--450,
  2010.

\bibitem{alves2021children}
P.~Alves-Oliveira, P.~Arriaga, A.~Paiva, and G.~Hoffman, ``Children as robot
  designers,'' in \emph{Proceedings of the 2021 ACM/IEEE International
  Conference on Human-Robot Interaction}, 2021, pp. 399--408.

\bibitem{winkle2020situ}
K.~Winkle, S.~Lemaignan, P.~Caleb-Solly, P.~Bremner, A.~Turton, and
  U.~Leonards, ``In-situ learning from a domain expert for real world socially
  assistive robot deployment,'' \emph{Proceedings of Robotics: Science and
  Systems. Corvalis, Oregon, USA.}, 2020.

\bibitem{senft2019teaching}
E.~Senft, S.~Lemaignan, P.~E. Baxter, M.~Bartlett, and T.~Belpaeme, ``Teaching
  robots social autonomy from in situ human guidance,'' \emph{Science
  Robotics}, vol.~4, no.~35, 2019.

\bibitem{winkle2020mutual}
K.~Winkle, P.~Caleb-Solly, A.~Turton, and P.~Bremner, ``Mutual shaping in the
  design of socially assistive robots: a case study on social robots for
  therapy,'' \emph{International Journal of Social Robotics}, vol.~12, no.~4,
  pp. 847--866, 2020.

\bibitem{bjorlingExploringTeensRobot2020}
E.~A. Bj{\"o}rling, K.~Thomas, E.~J. Rose, and M.~Cakmak, ``Exploring {{Teens}}
  as {{Robot Operators}}, {{Users}} and {{Witnesses}} in the {{Wild}},''
  \emph{Frontiers in Robotics and AI}, vol.~7, 2020.

\bibitem{ritchie2013qualitative}
J.~Ritchie, J.~Lewis, C.~M. Nicholls, R.~Ormston \emph{et~al.},
  \emph{Qualitative research practice: A guide for social science students and
  researchers}.\hskip 1em plus 0.5em minus 0.4em\relax sage, 2013.

\bibitem{tuckman1965developmental}
B.~W. Tuckman, ``Developmental sequence in small groups.'' \emph{Psychological
  bulletin}, vol.~63, no.~6, p. 384, 1965.

\end{thebibliography}
